\pdfoutput=1
\documentclass[english]{article}

\usepackage{algorithmic}
\usepackage{algorithm}
\usepackage{amsfonts}
\usepackage{amsmath}
\usepackage{amssymb}
\usepackage{amsthm}
\usepackage{array}
\usepackage{arydshln}
\usepackage{bm}
\usepackage{cite}
\usepackage{color}
\usepackage{comment}
\usepackage{dsfont}
\usepackage{enumitem}
\usepackage{float}
\usepackage[T1]{fontenc}
\usepackage{geometry}
\usepackage{graphicx}
\usepackage{grffile}
\usepackage{hyperref}\hypersetup{colorlinks,linkcolor=blue,anchorcolor=blue,citecolor=blue}
\usepackage[latin9]{inputenc}
\usepackage{letltxmacro}
\usepackage{mathrsfs}
\usepackage{mathtools}
\usepackage{multirow}
\usepackage{nicefrac}
\usepackage{subcaption}
\usepackage{tablefootnote}
\usepackage{verbatim}

\usepackage{todonotes}

\newcommand{\br}{\bm{r}}

\newcommand{\bF}{\bm{F}}

\newcommand{\bU}{\bm{U}}

\newcommand{\bX}{\bm{X}}

\newcommand{\cL}{\mathcal{L}}
\newcommand{\cM}{\mathcal{M}}

\newcommand{\cT}{\mathcal{T}}

\newcommand{\bcA}{\bm{\mathcal{A}}}
\newcommand{\bcB}{\bm{\mathcal{B}}}

\newcommand{\bcG}{\bm{\mathcal{G}}}

\newcommand{\bcM}{\bm{\mathcal{M}}}

\newcommand{\bcS}{\bm{\mathcal{S}}}

\newcommand{\bcX}{\bm{\mathcal{X}}}
\newcommand{\bcY}{\bm{\mathcal{Y}}}

\newcommand{\RR}{\mathbb{R}}

\newcommand{\Matricize}[2]{\mathcal{M}_{#1}\left(#2\right)}
\newcommand{\Shrink}[2]{\mathcal{T}_{#1}\left(#2\right)}
\newcommand{\hosvd}[2]{\mathcal{H}_{#1}\left(#2\right)}
\newcommand{\norm}[1]{\left\lVert#1\right\rVert}
\newcommand{\inner}[2]{\left\langle#1,#2\right\rangle}

\DeclareMathOperator{\bcdot}{\boldsymbol{\cdot}}

\DeclareMathOperator{\fro}{\mathsf{F}}

\allowdisplaybreaks
\makeatletter

\theoremstyle{plain} 

\theoremstyle{definition}
 
\theoremstyle{remark}

\definecolor{tian}{RGB}{0,150,0}
\definecolor{cm}{RGB}{250,0,200}
\definecolor{yc}{RGB}{255,0,0}
\definecolor{hd}{RGB}{0,180,200}

\begin{document}
\title{Deep Unfolded Tensor Robust PCA with Self-supervised Learning}
%
 
\author
 {
 	Harry Dong\thanks{Department of Electrical and Computer Engineering, Carnegie Mellon University, Pittsburgh, PA 15213, USA; Emails:
 		\texttt{\{harryd,yuejiec\}@andrew.cmu.edu}.
    } \\
	CMU \\
	\and
 	Megna Shah \thanks{Materials and Manufacturing Directorate, Air Force Research Laboratory, Wright-Patterson AFB, OH 45433, USA; Emails:
 		\texttt{\{megna.shah.1,sean.donegan\}@afrl.af.mil}.}\\
 		AFRL \\
 		\and
 	Sean Donegan\footnotemark[2]\\
 		 AFRL \\
		\and
 	Yuejie Chi\footnotemark[1] \\
 	CMU
 } 
 
\date{\today}

\setcounter{tocdepth}{2}
\maketitle

\begin{abstract}

Tensor robust principal component analysis (RPCA), which seeks to separate a low-rank tensor from its sparse corruptions, has been crucial in data science and machine learning where tensor structures are becoming more prevalent. While powerful, existing tensor RPCA algorithms can be difficult to use in practice, as their performance can be sensitive to the choice of additional hyperparameters, which are not straightforward to tune. In this paper, we describe a fast and simple self-supervised model for tensor RPCA using deep unfolding by only learning four hyperparameters. Despite its simplicity, our model expunges the need for ground truth labels while maintaining competitive or even greater performance compared to supervised deep unfolding. Furthermore, our model is capable of operating in extreme data-starved scenarios. We demonstrate these claims on a mix of synthetic data and real-world tasks, comparing performance against previously studied supervised deep unfolding methods and Bayesian optimization baselines.

\end{abstract}



\section{Introduction}
\label{sec:intro}

Becoming increasingly pervasive in data science and machine learning, tensors are powerful data structures that capture interactions between elements in higher dimensions. These structures lead to inherent properties that can be exploited for tasks that seek to analyze and process tensors. One such task is robust principal component analysis (RPCA), the problem of recovering a low-rank tensor that has been sparsely corrupted, which has found numerous applications such as surveillance, anomaly detection, and more.   


Compared with its matrix counterpart, dealing with tensor RPCA faces some unique challenges, as many ideas from matrix RPCA begin to fall apart. First, blindly applying a matrix RPCA algorithm to a flattened tensor can destroy structural information \cite{yuan2016tensor}. Second, convex formulations, typically done with the nuclear norm as a convex relaxation to the rank constraint, is NP-hard for tensors \cite{friedland2018nuclear}. Third, there are multiple notions of tensor decomposition and rank. As such, provable tensor RPCA algorithms have been developed for low multilinear rank \cite{dong2022fast}, tubal rank \cite{lu2016tensor, lu2019tensor}, and CP-rank \cite{driggs2019tensor, anandkumar2016tensor} decompositions. 

In this paper, we focus on the Tucker decomposition with low multilinear rank, though we believe our approach is also extendable to other tensor decompositions.
While powerful, existing efficient tensor RPCA algorithms, e.g. based on iterative scaled gradient updates as in \cite{dong2022fast}, can be difficult to use in practice, as their performance is sensitive to the choice of hyperparameters, such as learning rates and thresholds. 
In addition, tuning these hyperparameters can be difficult, since the number of hyperparameters scales with the desired number of iterations. 

Leveraging deep unfolding, the process of unrolling an iterative algorithm into a deep neural network introduced in \cite{gregor2010learning}, there have been considerable efforts in learning algorithmic parameters  with backpropagation to improve the performance of the original algorithm. Deep unfolding for RPCA has been applied to ultrasound imaging \cite{solomon2019deep, cohen2019deep}, background subtraction \cite{van2021deep}, and the special case of positive semidefinite (PSD) low-rank matrices \cite{herrera2020denise}. To generalize this idea, \cite{cai2021learned} designed a deep unfolded architecture for matrix RPCA, which is extendable to infinitely many RPCA iterations. However, many existing approaches \cite{solomon2019deep, cohen2019deep, van2021deep, cai2021learned} train on ground truth labels which may be limited or nonexistent in practice. Though only applied to PSD low-rank matrices, \cite{herrera2020denise} addressed this issue with an unsupervised model. 

Motivated by the need to broaden the applicability and improve the performance of tensor RPCA in practice, we describe a
fast and simple self-supervised model for tensor RPCA by  unfolding the recently proposed algorithm in \cite{dong2022fast}, due to its appealing performance. Our contributions are as follows.
\begin{enumerate}
    \item We propose a novel self-supervised model for tensor RPCA, which can be used independently or as fine tuning for its supervised counterpart. Furthermore, it scales to an arbitrary number of RPCA iterative updates with only four learnable parameters by leveraging theoretical insights from \cite{dong2022fast}.
    \item Synthetic and real-world experiments on video surveillance and materials microscopy data show that our self-supervised model matches or exceeds the performance of supervised learning without the need for ground truth, especially in data-starved scenarios.
\end{enumerate}

 \paragraph{Paper organization.}
The rest of this paper is organized as follows. We build up the tensor RPCA algorithm from \cite{dong2022fast} in Section \ref{sec:background} which will be used in our learned tensor RPCA method described in Section \ref{sec:model}. Then, we present synthetic and real-world experimental results in Section \ref{sec:experiments}, followed by final remarks in Section \ref{sec:conclusion}. 

\paragraph{Notation and basics of tensor algebra.} Throughout this paper, we represent tensors with bold calligraphic letters (e.g. $\bcX$) and matrices with bold capitalize letters (e.g. $\bX$). For tensor $\bcA  \in \RR^{n_1 \times n_2 \times n_3}$,  $\Matricize{k}{\bcA}$ is the tensor matricization along the $k$-th mode (i.e. dimension of the tensor), $k=1,2,3$, and $[\bcA]_{i_1, i_2, i_3}$ is the $(i_1, i_2, i_3)$-th entry of $\bcA$. A fiber is a vector obtained by fixing all indices except one in the tensor (e.g. $[\bcA]_{i_1, \cdot, i_3}$). Let the inner product between two tensors $\bcA$ and $\bcB$ be $\inner{\bcA}{\bcB} = \sum_{i_1, i_2, i_3} [\bcA]_{i_1, i_2, i_3} [\bcB]_{i_1, i_2, i_3}$, and $\norm{\bcA}_{\fro} = \sqrt{\inner{\bcA}{\bcA}}$ and $\norm{\bcA}_{1} = \sum_{i_1, i_2, i_3} |[\bcA]_{i_1, i_2, i_3}|$ denote the Frobenius norm and the $\ell_1$-norm, respectively. Furthermore, suppose a tensor $\bcX \in \RR^{n_1 \times n_2 \times n_3}$ has multilinear rank $\br = (r_1, r_2, r_3)$, and its Tucker decomposition is given by $\bcX = (\bU^{(1)}, \bU^{(2)}, \bU^{(3)}) \bcdot \bcG$ with $\bU^{(k)} \in \RR^{n_k \times r_k}$ and $\bcG \in \RR^{r_1 \times r_2 \times r_3}$, it follows that
\begin{align*}
    [\bcX]_{i_1, i_2, i_3} = \sum_{j_1, j_2, j_3} \Bigg(\prod_{k=1, 2, 3} [\bU^{(k)}]_{i_k, j_k} \Bigg)[\bcG]_{j_1, j_2, j_3} .
\end{align*}
Last but not least, given \textit{any} $\bcX \in \RR^{n_1 \times n_2 \times n_3}$, its rank-$\br$ higher-order singular value decomposition (HOSVD) $ \hosvd{\br}{\bcX}$ is given by
$\hosvd{\br}{\bcX} = (\bU^{(1)}, \bU^{(2)}, \bU^{(3)}, \bcG)$,
where $\bU^{(k)} \in \RR^{n_k \times r_k}$ contains the top $r_k$ singular vectors of $\cM_k(\bcX)$ and $\bcG = (\bU^{(1) \top}, \bU^{(2) \top}, \bU^{(3) \top}) \bcdot \bcX$.

\section{Background on Tensor RPCA}
\label{sec:background}

\subsection{Problem formulation}
\label{ssec:formulation}


Suppose we observe a corrupted tensor $\bcY\in \RR^{n_1 \times n_2 \times n_3}$ of the form
\begin{align}
    \bcY = \bcX_\star + \bcS_\star ,
\end{align}
where $\bcX_\star$ is a low-rank tensor with the multilinear rank $\br = (r_1, r_2, r_3)$, whose Tucker decomposition is given by
$\bcX_\star = (\bU_\star^{(1)}, \bU_\star^{(2)}, \bU_\star^{(3)}) \bcdot \bcG_\star$ with $\bU_\star^{(k)} \in \RR^{n_k \times r_k}$ for $k=1, 2, 3$, and $\bcG_\star \in \RR^{r_1 \times r_2 \times r_3}$. Moreover, $\bcS_\star$ is an $\alpha$-sparse corruption tensor, i.e. $\bcS_\star$ contains at most $0 \leq \alpha < 1$ fraction nonzero values along each fiber. Given $\bcY$ and $\br$, the goal of tensor RPCA aims to accurately obtain $\bcX_\star$ and $\bcS_\star$.


\subsection{Tensor RPCA via ScaledGD}
\label{ssec:scaledgd}

Recently, Dong et al. \cite{dong2022fast} proposed a fast and scalable method for tensor RPCA with provable performance guarantees. Specifically, they try to minimize the objective function
\begin{align}\label{eq:loss}
    \cL(\bF,\bcS) \coloneqq \frac{1}{2}\left\| \big(\bU^{(1)},\bU^{(2)},\bU^{(3)} \big)\bcdot\bcG+\bcS-\bcY \right\|_{\fro}^{2}, 
\end{align}
where $\bF = (\bU^{(1)}, \bU^{(2)}, \bU^{(3)}, \bcG)$ and $\bcS$ are the estimates of the factors of the low-rank tensor and the sparse tensor, respectively. The algorithm proceeds by iterative updates of the factors $\bF$ via scaled gradient descent (ScaledGD) \cite{tong2021accelerating,tong2022scaling,tong2022accelerating,tong2021low}, followed by filtering the outliers $\bcS$ using the shrinkage operator $\cT_{\zeta}(\cdot)$, defined as
\begin{align}
    [\cT_{\zeta}(\bcX)]_{i, j, k} \coloneqq \mathrm{sgn}([\bcX]_{i, j, k}) \cdot \max(0, |[\bcX]_{i, j, k}| - \zeta).
\end{align}
The details of the proposed ScaledGD algorithm are summarized in Algorithm \ref{alg:tensor_RPCA}. Beginning with a careful initialization using the spectral method,  given by
\begin{align}
    \big(\bU_0^{(1)}, \bU_0^{(2)}, \bU_0^{(3)}, \bcG_0 \big) = \hosvd{\br}{\bcY - \Shrink{\zeta_0}{\bcY}},
\end{align}
for each iteration, ScaledGD makes the following updates:
\begin{subequations} \label{eq:alg1_updates}
\begin{align} 
    \bcS_{t+1} &= \Shrink{\zeta_{t+1}}{\bcY - \big(\bU_t^{(1)}, \bU_t^{(2)}, \bU_t^{(3)} \big) \bcdot \bcG_t},  \label{eq:update_corruption} \\
    \bU^{(k)}_{t+1} &= \bU^{(k)}_t - \eta \nabla_{\bU^{(k)}_t} \cL(\bF_t, \bcS_{t+1}) \big(\Breve{\bU}_t^{(k)\top} \Breve{\bU}_t^{(k)} \big)^{-1} \label{a_update3}
\end{align}
for $k=1,2,3$, and
\begin{align}
    \bcG_{t+1} &= \bcG_t - \eta \left(\Tilde{\bU}^{(1)}_t, \Tilde{\bU}^{(2)}_t, \Tilde{\bU}^{(3)}_t \right) \bcdot \nabla_{\bcG_t} \cL(\bF_t , \bcS_{t+1}) ,\label{g_update3}
\end{align}
\end{subequations}
where $\eta >0$ is the learning rate, and  $\{\zeta_t\}_{t=0}^{T}$ is the threshold schedule up to $T$ iterations. Here,
\begin{align*}
    \Breve{\bU}_t^{(1)} &= (\bU_t^{(3)} \otimes \bU_t^{(2)}) \cM_1(\bcG_t)^\top, \\
    \Breve{\bU}_t^{(2)} &= (\bU_t^{(3)} \otimes \bU_t^{(1)}) \cM_2(\bcG_t)^\top, \\
    \Breve{\bU}_t^{(3)} &= (\bU_t^{(2)} \otimes \bU_t^{(1)}) \cM_3(\bcG_t)^\top,
\end{align*}
and $\Tilde{\bU}^{(k)}_t = \big(\bU^{(k)\top}_t \bU_t^{(k)} \big)^{-1}$, $k=1,2,3$, where $\otimes$ denotes the Kronecker product. In \cite{dong2022fast}, it is demonstrated that ScaledGD allows perfect recovery of the low-rank tensor $\bcX_{\star}$ and the sparse tensor $\bcS_{\star}$ under mild assumptions as long as the low-rank tensor is incoherent and the corruption level $\alpha$ is not too large. Moreover, ScaledGD converges at a linear rate independent of the condition number of $\bcX_\star$ \cite{dong2022fast}, making it more appealing than the vanilla gradient descent approach that is much more sensitive to ill-conditioning. However, translating the theoretical advantage into practice requires carefully tuned hyperparamters, $\eta$ and $\{\zeta_t\}_{t=0}^{T}$, which unfortunately are not readily available.


\begin{algorithm}[t]
\caption{ScaledGD for 3rd order tensor RPCA}\label{alg:tensor_RPCA} 
\begin{algorithmic} 
\STATE \textbf{Input:} the observed tensor $\bcY$, the multilinear rank $\br$, learning rate $\eta$, and threshold schedule $\{\zeta_t\}_{t=0}^{T}$. 
\STATE \textbf{Init:} $\big(\bU_0^{(1)}, \bU_0^{(2)}, \bU_0^{(3)}, \bcG_0 \big) = \hosvd{\br}{\bcY - \Shrink{\zeta_0}{\bcY}}$.
\FOR{$t = 0, 1, \dots, T-1$}
    \STATE Update $\bcS_{t+1}$ via \eqref{eq:update_corruption};
    \STATE Update $\bF_{t+1} = \big( \bU^{(1)}_{t+1},  \bU^{(2)}_{t+1},  \bU^{(3)}_{t+1},  \bcG_{t+1} \big)$ via \eqref{a_update3} and \eqref{g_update3};
\ENDFOR
\STATE \textbf{Output:} $ \bF_T = \big (\bU_{T}^{(1)}, \bU_T^{(2)}, \bU_T^{(3)}, \bcG_T \big) $.
\end{algorithmic} 
\end{algorithm}

\section{Proposed Method}
\label{sec:model}

We aim to greatly broaden the applicability of Algorithm \ref{alg:tensor_RPCA} with learned hyperparameters by leveraging self-supervised learning (SSL) under a deep-unfolding perspective \cite{gregor2010learning} of Algorithm \ref{alg:tensor_RPCA}.

\subsection{Unrolling-aided hyperparameter tuning}
\label{ssec:loss_func}

\begin{figure}[ht]
\begin{minipage}[b]{1\linewidth}
  \centering
  \includegraphics[width=0.7\linewidth]{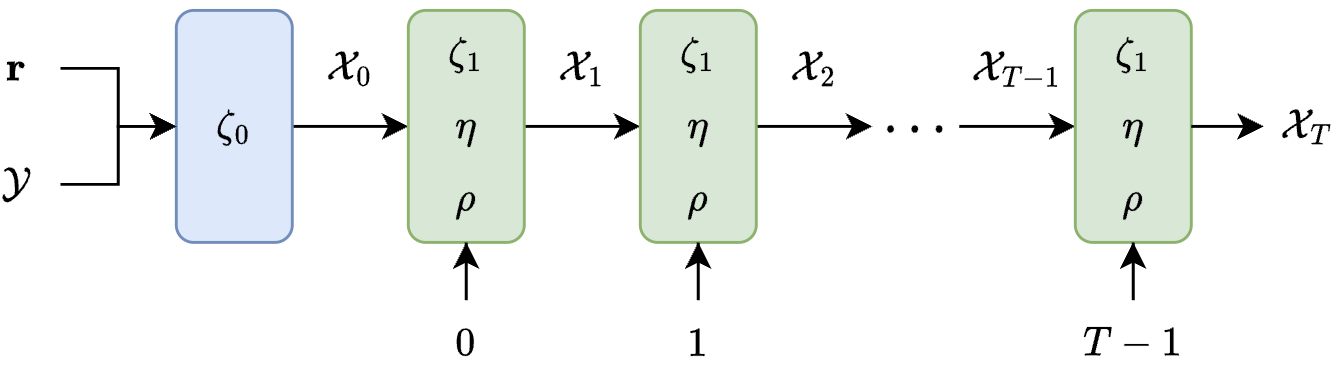}
\end{minipage}
\caption{Network architecture. $\bcY$ and $\br$ are fed into the first layer, the initialization step in Algorithm \ref{alg:tensor_RPCA}. Subsequent recurrent layers follow Algorithm \ref{alg:tensor_RPCA} to update $\bcX_t = (\bU_{t}^{(1)}, \bU_t^{(2)}, \bU_t^{(3)}) \bcdot \bcG_t$. Recurrent layers require the iteration count $t$ due to our definition of the $(t+1)$-th thresholding parameter, $\zeta_{t+1} \coloneqq  \zeta_1 \rho^{t}$ for $t \geq 0$. We use softplus activations to ensure $\zeta_0, \zeta_1, \eta > 0$ and the sigmoid function to make $0 < \rho< 1$. }
\label{fig:architecture}
\end{figure}

Based on the insights from theoretical analysis \cite[Theorem 1]{dong2022fast}, we use a threshold schedule that decays at every iteration to reduce the number of hyperparameters. In other words, 
$$\zeta_{t+1} = \rho \zeta_{t}$$ for $t \geq 1$ and $0 < \rho < 1$, so there are only 4 hyperparameters that need tuning: $\zeta_0$, $\zeta_1$, $\rho$, and $\eta$. By unrolling Algorithm \ref{alg:tensor_RPCA}, we obtain an unfolded version of Algorithm \ref{alg:tensor_RPCA}, where $\zeta_0$, $\zeta_1$, $\rho$, and $\eta$ are learnable; see Fig. \ref{fig:architecture} for more details. The model consists of a feed-forward layer and a recurrent layer to model the initialization and iterative steps of Algorithm \ref{alg:tensor_RPCA}. 

It remains to specify the loss function used to learn the hyperparameters. One immediate idea is to optimize the performance in reconstructing the true tensor $\bcX_{\star}$ by minimizing
\begin{align}\label{eq:rel_error}
    \cL_{\text{SL}}(\bF) \coloneqq \frac{\left\| \bcX_\star - \big(\bU^{(1)},\bU^{(2)},\bU^{(3)} \big)\bcdot\bcG\right\|_{\fro}^2}{\norm{\bcX_\star}_{\fro}^2},
\end{align}
which we refer to as supervised learning (SL) similar to \cite{cai2021learned}. While this is possible when trained on synthetic data---where we have access to the ground truth---it may not be applicable in general in the absence of ground truth.

One might naturally wonder if \eqref{eq:loss} can be used to optimize the hyperparameters.
Unfortunately, with the decaying threshold schedule, \eqref{eq:loss} can be trivially solved as we increase the number of iterations $T$ or choosing $\rho$ close to $0$, since the threshold $\zeta_t$ will rapidly approach $0$, making $\bcS_t \approx \bcY - \bcX_t$ by \eqref{eq:update_corruption}. Hence, the objective function \eqref{eq:loss} is no longer appropriate in learning the hyperparameters. As such, we will instead use the following loss function:
\begin{align}\label{eq:l1loss}
    \cL_{\text{SSL}}(\bF) \coloneqq \frac{1}{\norm{\bcY}_{\fro}^2}\left\| \bcY - \big(\bU^{(1)},\bU^{(2)},\bU^{(3)} \big)\bcdot\bcG\right\|_{1}
\end{align}
to encourage sparsity of the corruption tensor \cite{charisopoulos2021low, herrera2020denise}. We refer to this loss function as the self-supervised learning (SSL) loss since it does not require the ground truth and therefore, more amenable to real-data scenarios.  

\subsection{From supervised learning to self-supervised learning}
\label{ssec:training}

Depending on the loss function, our model can be used for both supervised or self-supervised learning of the hyperparameters. For supervised learning, suppose we have access to $\bcX_\star$ for all tensor RPCA problem instances in a training dataset, then the hyperparameters can be tuned by optimizing \eqref{eq:rel_error}. 


A drawback of a purely supervised approach is the assumption of a ``one size fits all'' set of hyperparameters that can be applied to other similar tensors. In fact, tensor inputs into RPCA are exceptionally complex and in general, cannot be adequately described in a few features such as their sparsity levels, ranks, and condition numbers. Fortunately, the tuned hyperparameters obtained from training serve as a warm start for further refinements. Similar to \cite{herrera2020denise}, we first learn a fixed set of hyperparameters on the training dataset (when available), which will be fine tuned for each tensor during test time by minimizing \eqref{eq:l1loss} individually for each tensor, leading to a self-supervised learning paradigm.




\section{Experiments}
\label{sec:experiments}

We conduct synthetic and real-world data experiments to corroborate the effectiveness of the proposed learned tensor RPCA approach. Code available at 
\begin{center}
\url{https://github.com/hdong920/Tensor_RPCA_ScaledGD}.
\end{center} 

\subsection{Synthetic data}
\label{sec:synthetic_data}

We explore the effect of fine tuning with SSL for each combination of $\alpha \in \{0, 0.1, \dots, 0.9\}$ and $r \in \{10, 20, \dots, 70\}$ with fixed $n = 100$. When a rank $(r, r, r)$ tensor $\bcY \in \RR^{n \times n \times n}$ needs to be sampled, we first randomly generate factor matrices $\bU_\star^{(1)}, \bU_\star^{(2)}, \bU_\star^{(3)} \in \RR^{n \times r}$ with orthonormal columns and core tensor $\bcG_\star$ where $[\bcG_\star]_{i, i, i} = \kappa^{-(i-1)/(r-1)}$ for $\kappa = 5$ and $0$ elsewhere to construct $\bcX_\star = (\bU_\star^{(1)}, \bU_\star^{(2)}, \bU_\star^{(3)}) \bcdot \bcG_\star$. In addition, $\bcS_\star$ is a sparse tensor with $\alpha$-fraction of its entries drawn from a uniform distribution in $(-\theta, \theta)$, where $\theta \coloneqq  \frac{1}{n^3} \norm{\bcX_\star}_1$. All experiments in this section were run for $T=100$ iterations of Algorithm \ref{alg:tensor_RPCA}.

First, we perform supervised learning by minimizing \eqref{eq:rel_error}
similar to \cite{cai2021learned} for each combination of $\alpha$ and $r$. For each gradient step, we generate a new sample (which is a tensor RPCA problem instance). We train for 1000 gradient steps with a decaying learning rate scheduler. Next, we perform SSL for 500 gradient steps to fine tune the hyperparameters for 20 samples generated for each combination of $\alpha$ and $r$ by minimizing \eqref{eq:l1loss}.
As the baseline, we use Optuna \cite{optuna_2019}, a Bayesian optimization-based hyperparameter tuning package, to minimize \eqref{eq:l1loss} individually for 20 samples. We run Optuna for 500 iterations for each sample over a search space wide enough to contain the learned hyperparameters.
\begin{figure}[ht]
    \begin{minipage}[b]{1\linewidth}
     \centering
     \begin{subfigure}[b]{.34\linewidth}
         \centering
		 \includegraphics[width=0.87\linewidth]{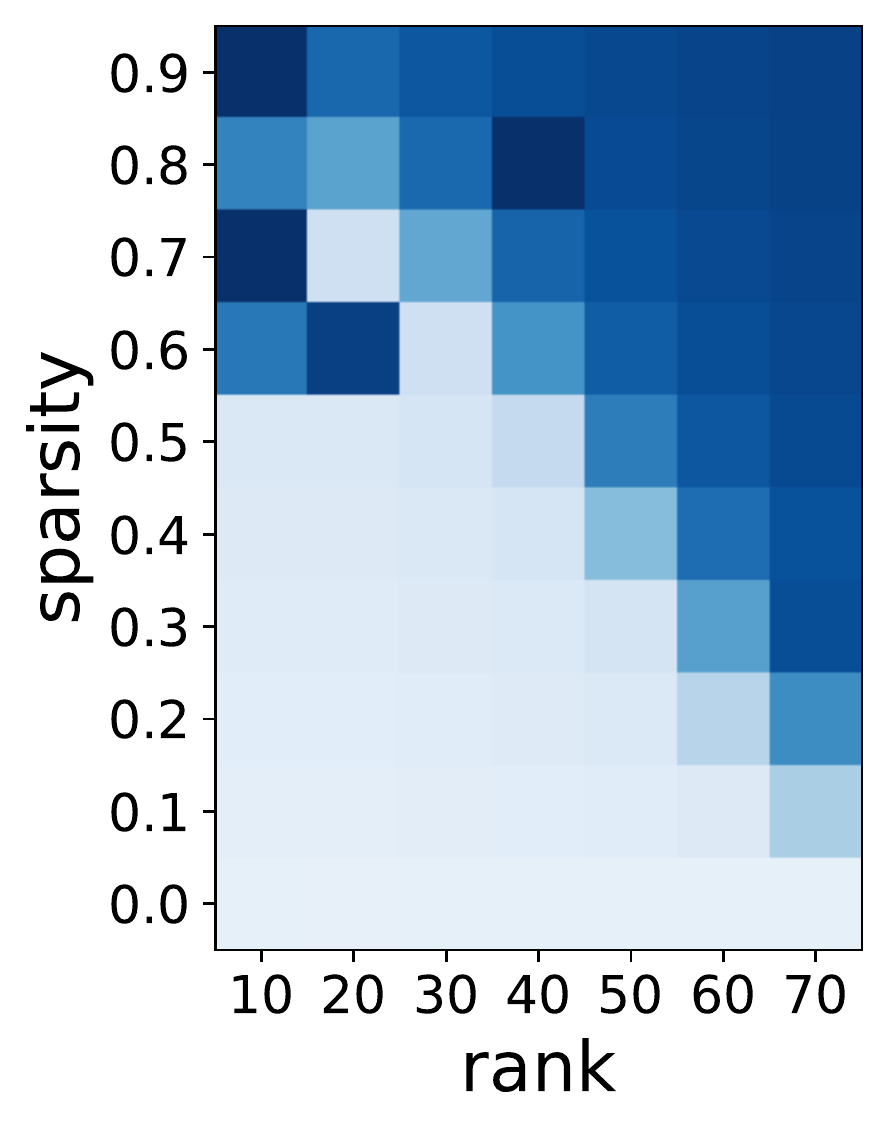}
         \caption{Baseline}
         \label{fig:numerical_baseline}
     \end{subfigure}
     \begin{subfigure}[b]{.3\linewidth}
         \centering
		 \includegraphics[width=0.793\linewidth]{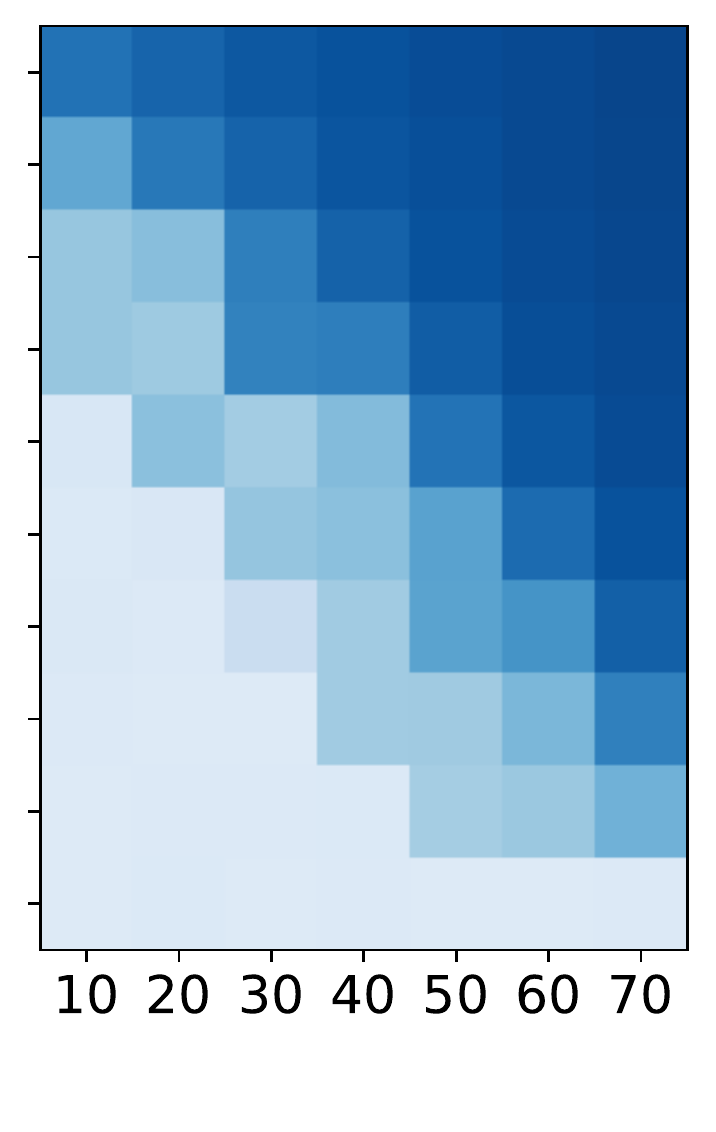}
         \caption{Supervised}
         \label{fig:numerical_supervised}
     \end{subfigure}
     \begin{subfigure}[b]{.34\linewidth}
         \centering
         \includegraphics[width=0.908\linewidth]{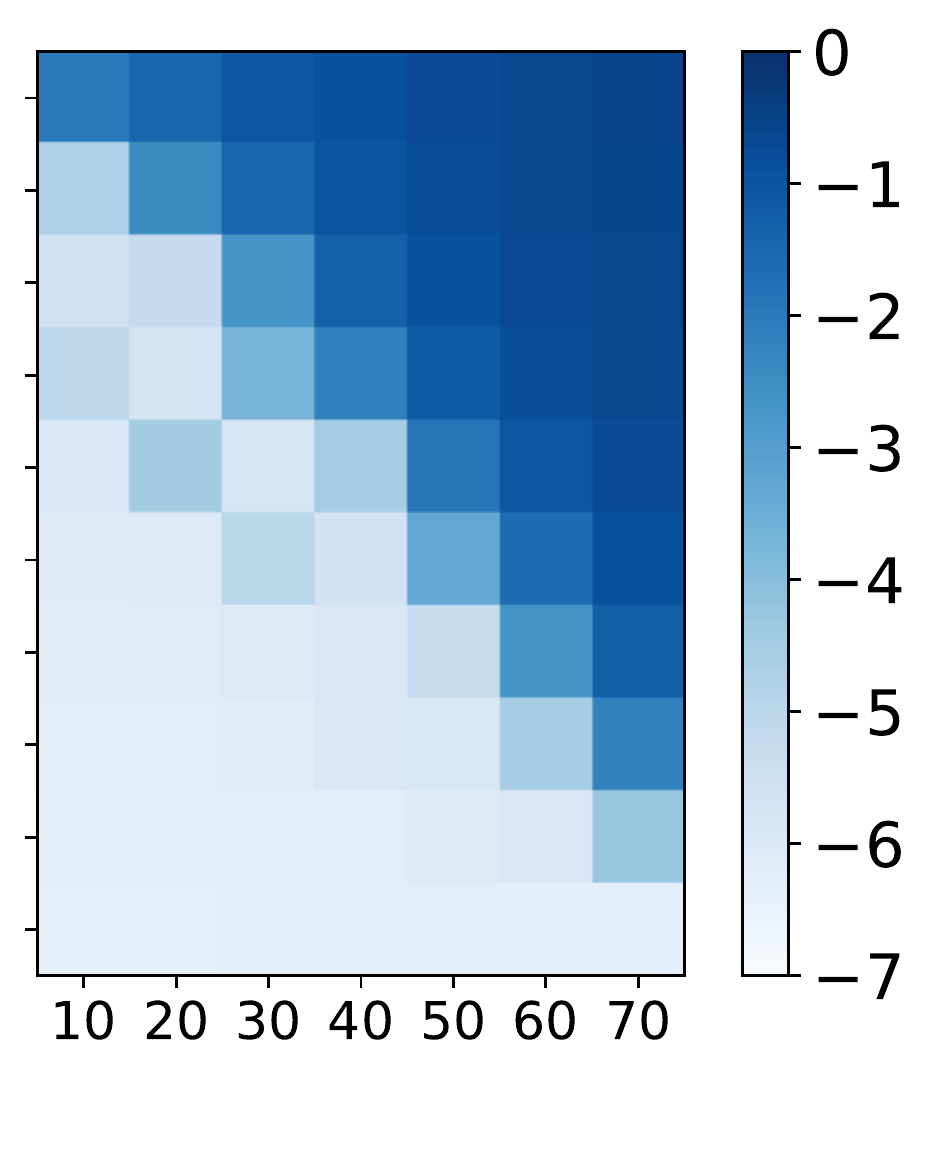}
         \caption{Fine tuning}
         \label{fig:numerical_fine_tuning}
     \end{subfigure}
     \end{minipage}
        \caption{The log mean relative recovery error, $\frac{\norm{\bcX_\star - \bcX_T}_{\fro}}{\norm{\bcX_\star}_{\fro}}$ for each combination of $\alpha$ and $r$ of Algorithm~\ref{alg:tensor_RPCA} using hyperparameters found by the baseline, supervised learning, and supervised learning with self-supervised fine tuning.}
        \label{fig:comparison}
\end{figure}
From Fig. \ref{fig:comparison}, we observe that the baseline accurately recovers $\bcX_\star$ across a wide range of cases, though it can be unstable in low rank and high sparsity scenarios. Supervised learning improves the stability of these scenarios but performs slightly worse along the edge of the phase transition. Although fine tuning adds more computation, we simultaneously obtain more stable results than the baseline and gain a significant improvement on recovering $\bcX_\star$ in a broad set of scenarios over supervised learning, especially along the edge of the phase transition.


During self-supervised fine tuning, we track the percent change of the RPCA hyperparameters' fined tuned values relative to their final values from supervised learning. Table \ref{tab:parameter_change} contains the quartiles of these percent changes  for samples that observed a more than a 90\% reduction in recovery error after fine tuning. From this, we see that RPCA hyperparameters can be quite sensitive where tuning by grid search would require very fine sampling. Fine tuning adapts to the sensitivity of RPCA hyperparameters which supervised learning alone cannot address.
\begin{table}[ht]
    \centering
    \vspace{-0.0in}
    \begin{tabular}{ |p{1.3cm}||p{1.3cm}|p{1.3cm}|p{1.3cm}|p{1.3cm}|  }
     \hline
     Quartile & $\zeta_0$ &$\zeta_1$ &$\eta$ &$\rho$\\
     \hline
     Q1 &0.93\% &3.06\% &-14.03\% &3.36\% \\
     Q2 &5.17\% &28.91\% &-3.38\% &6.20\% \\
     Q3 &15.73\% &36.56\% &7.85\% &6.68\% \\
     \hline
    \end{tabular}
    \caption{Quartiles of fine tuning parameter percent change for samples with greater than $90\%$ reduction in recovery loss after fine tuning.}
    \label{tab:parameter_change}
\end{table}


\subsection{Background subtraction in video surveillance}
\label{sec:background_subtraction}

\begin{figure}[ht]
    \begin{minipage}[b]{1\linewidth}
     \centering
     \begin{subfigure}[b]{0.48\linewidth}
         \centering
		 \includegraphics[width=0.95\linewidth]{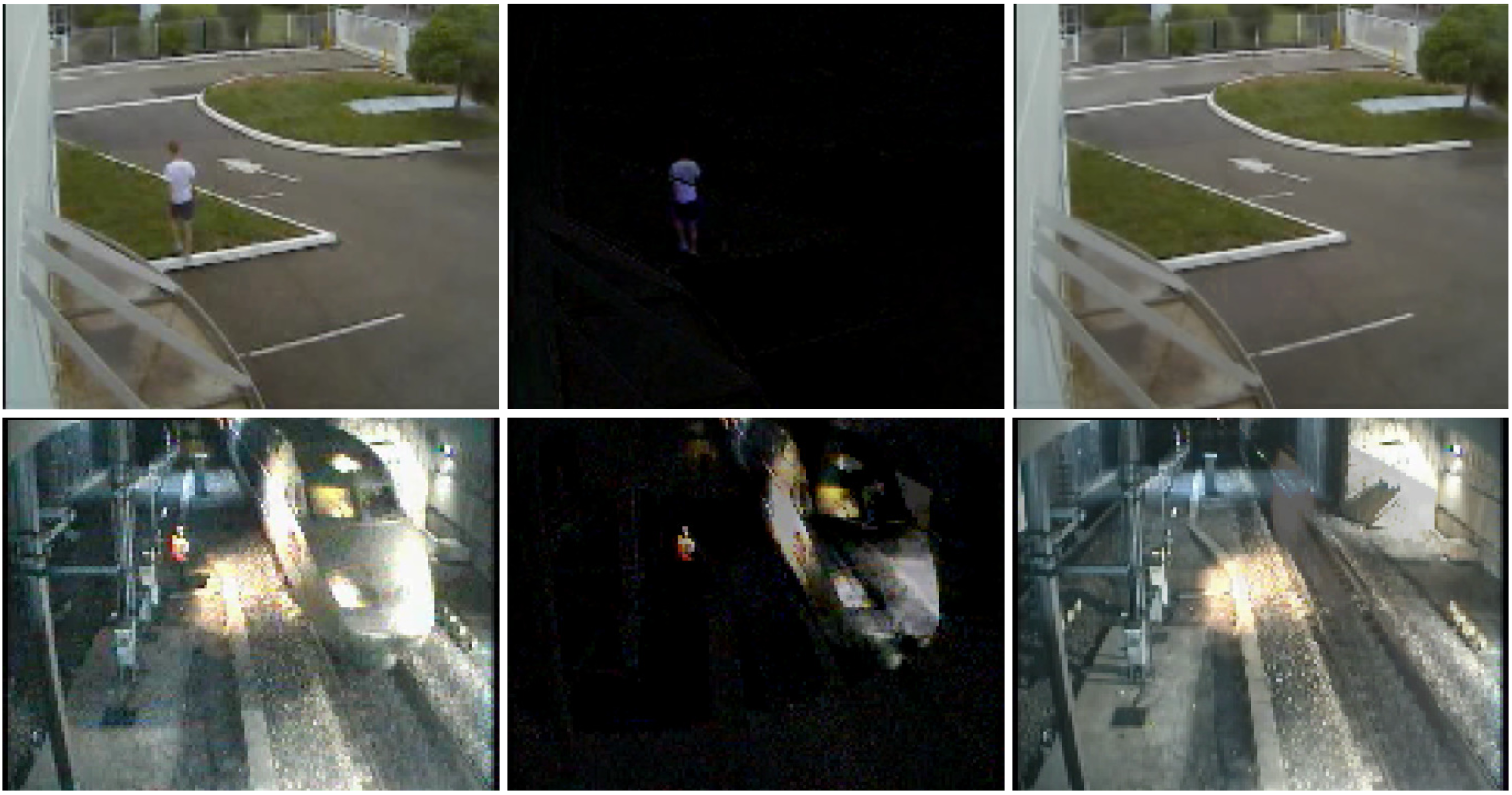}\\
         \caption{Supervised learning}
         \label{fig:background_supervised}
     \end{subfigure}
     \begin{subfigure}[b]{0.48\linewidth}
         \centering
         \includegraphics[width=0.95\linewidth]{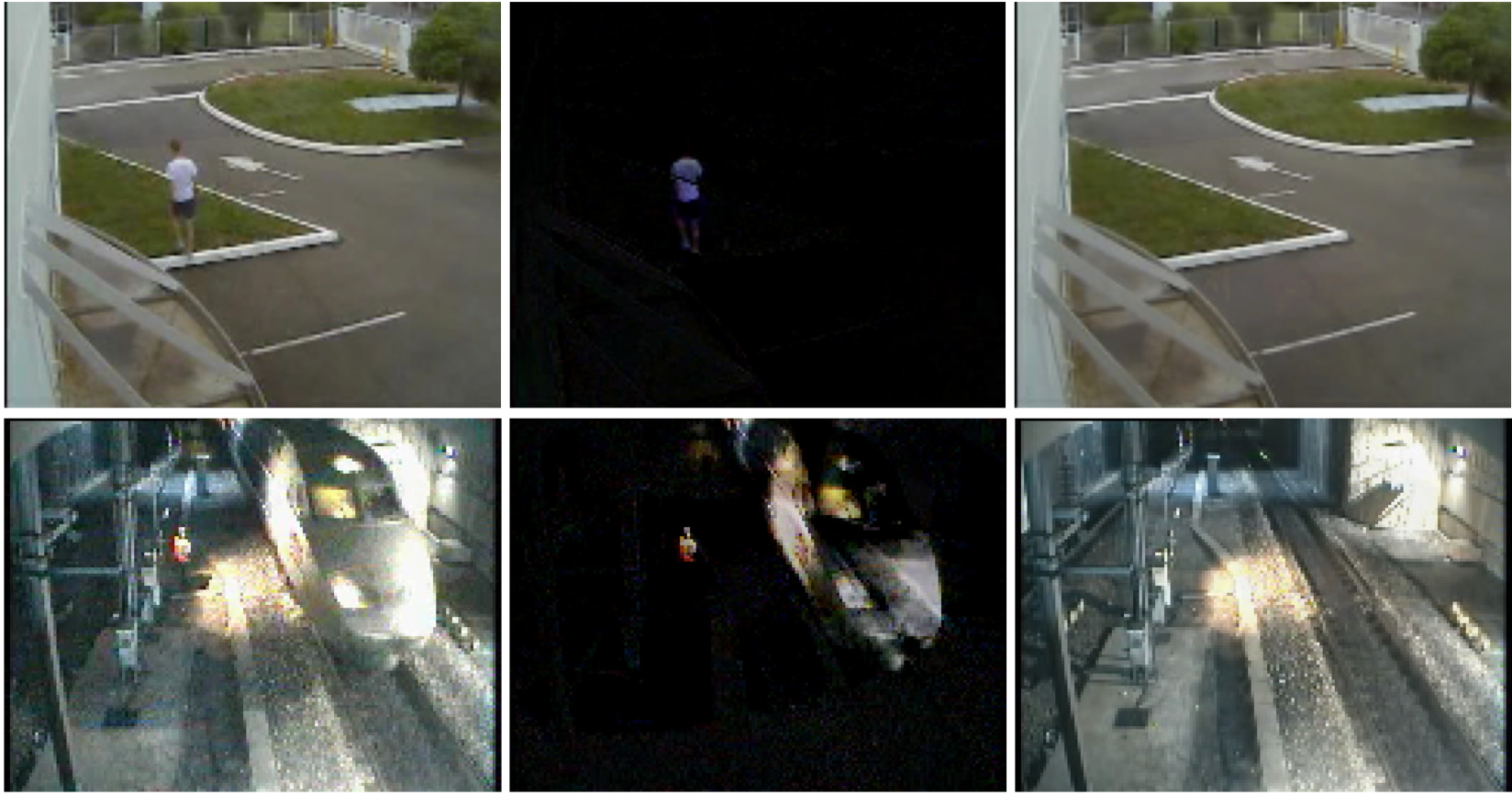} \\
         \caption{Self-supervised learning}
         \label{fig:background_ssl}
     \end{subfigure}
     \end{minipage}
        \caption{Background subtraction of a video containing a moving train and worker. From left to right of each half, the columns are the input video, extracted sparse foreground, and low-rank background. }
        \label{fig:background_subtraction}
\end{figure}

Next, we apply learned tensor RPCA from scratch on the background subtraction task using the Background Models Challenge real videos dataset \cite{vacavant2012benchmark}, which contains 9 videos of varying shapes. We use the first 6 to train two separate models via supervised learning and SSL with $T=150$ for 15 epochs, which are tested on the last 3 videos. Models assumed rank 1 along the time dimension and full rank for all others. For SSL, it minimizes \eqref{eq:l1loss}. However, for supervised learning, as this dataset includes only ground truth binary foreground masks (Boolean labels that indicate if a pixel is part of the foreground), we cannot use \eqref{eq:rel_error} directly. Defining $\bcM$ to be the binary foreground mask of the same shape as $\bcY$, we know that $\bcS_\star$ and $\bcM$ share the same support. Letting $\odot$ be the Hadamard product, we use
\begin{align*}\label{eq:rel_error_mask}
    \cL_{\text{SM}}(\bF) \coloneqq \frac{\left\| \left(\bcY - \big(\bU^{(1)},\bU^{(2)},\bU^{(3)} \big)\bcdot\bcG \right) \odot (1 - \bcM) \right\|_{\fro}^2}{\norm{\bcY \odot (1 - \bcM)}_{\fro}^2} 
\end{align*}
for supervised learning since $\bcY \odot (1 - \bcM) = \bcX_\star \odot (1 - \bcM)$. Because we only assume low rank along the time dimension, iterative updates of Algorithm \ref{alg:tensor_RPCA} are skipped for other modes to save computation. Fig.~ \ref{fig:background_subtraction} display no significant visual difference between results from the two models, suggesting the benefit of SSL in alleviating the need of labeled training data for this task. This is backed up by their similar masked average relative recovery test errors, $\sqrt{\cL_{\text{SM}}(\bF_T)}$. We observe that as the assumed time dimension rank increased, the more fine tuning improved relative recovery error.

\subsection{Electron backscatter diffraction microscopy}
\label{sec:ebsd}

\begin{figure*}[!t]
    \begin{minipage}[b]{1\linewidth}
     \centering
     \begin{subfigure}[b]{1\linewidth}
         \centering
         \includegraphics[width=0.75\linewidth]{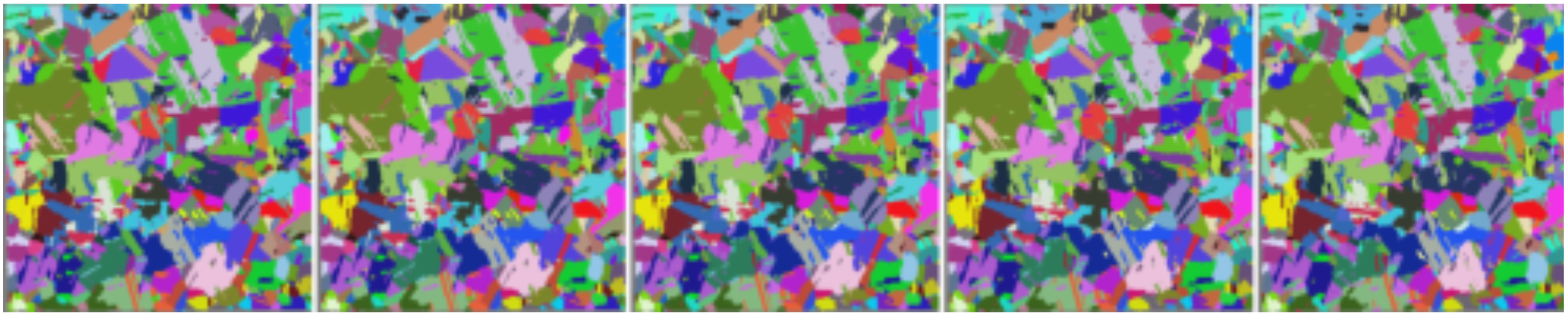} \\
         \caption{Original slices}
         \label{fig:ebsd_y}
     \end{subfigure}
     \begin{subfigure}[b]{1\linewidth}
         \centering
         \includegraphics[width=0.75\linewidth]{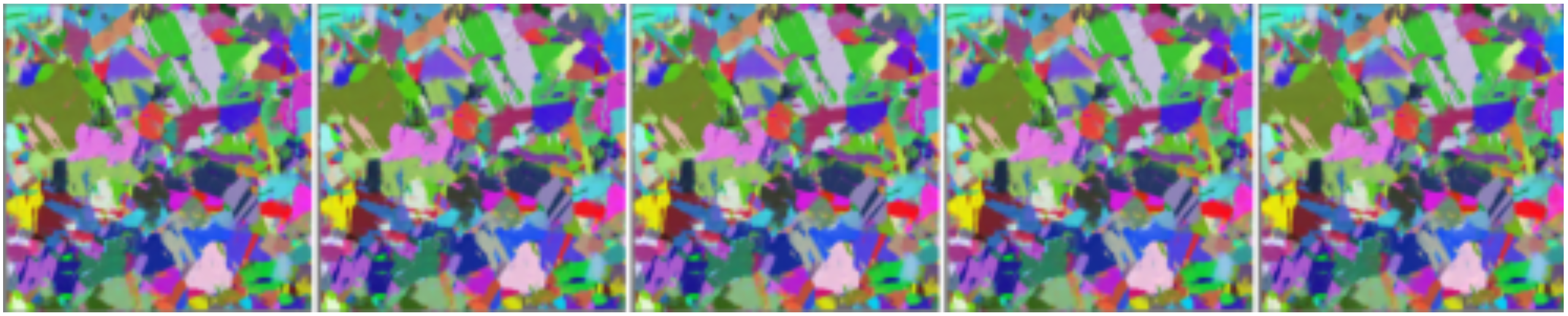} \\
         \caption{Low rank component produced by SSL}
         \label{fig:ebsd_ssl_x}
     \end{subfigure}
     \begin{subfigure}[b]{1\linewidth}
         \centering
         \includegraphics[width=0.75\linewidth]{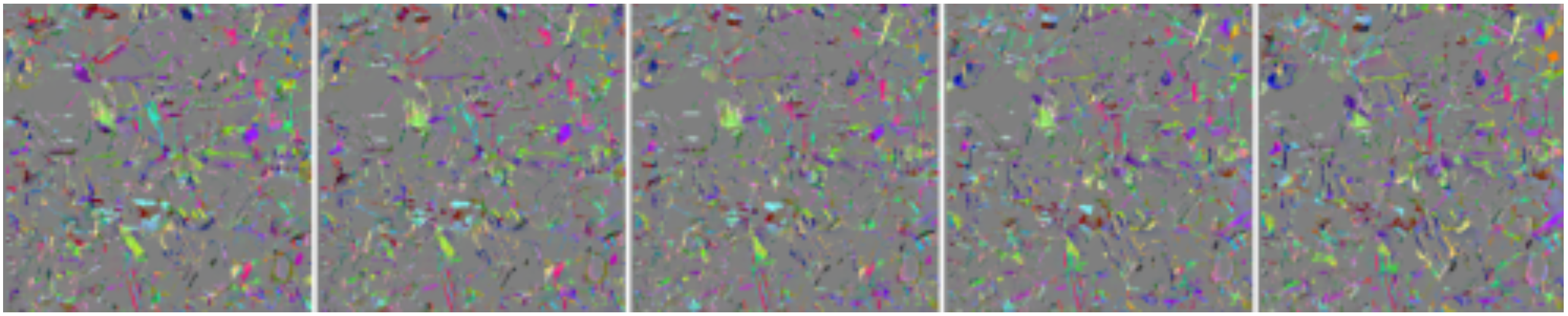} \\
         \caption{Sparse component produced by SSL}
         \label{fig:ebsd_ssl_s}
     \end{subfigure}
     \begin{subfigure}[b]{1\linewidth}
         \centering
         \includegraphics[width=0.75\linewidth]{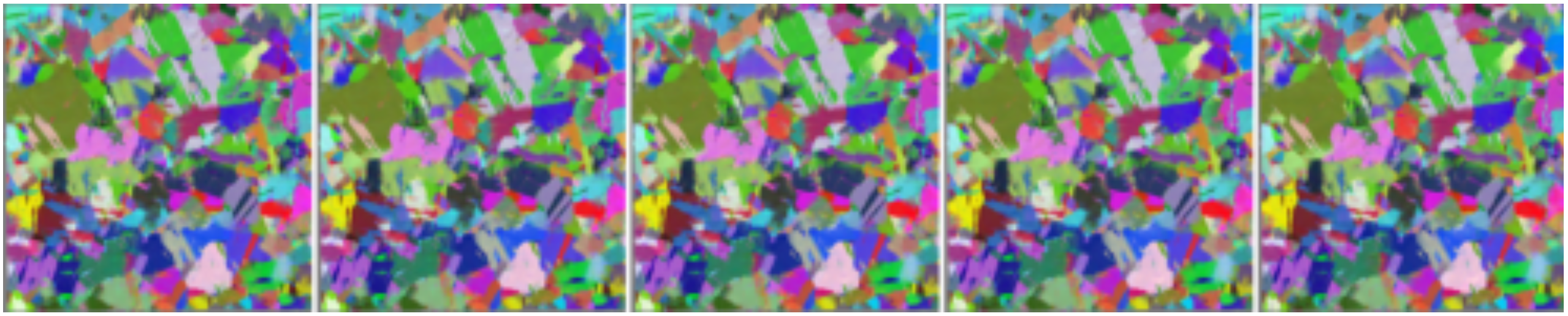} \\
         \caption{Low rank component produced by Baseline}
         \label{fig:ebsd_bayesian_x}
     \end{subfigure}
     \begin{subfigure}[b]{1\linewidth}
         \centering
         \includegraphics[width=0.75\linewidth]{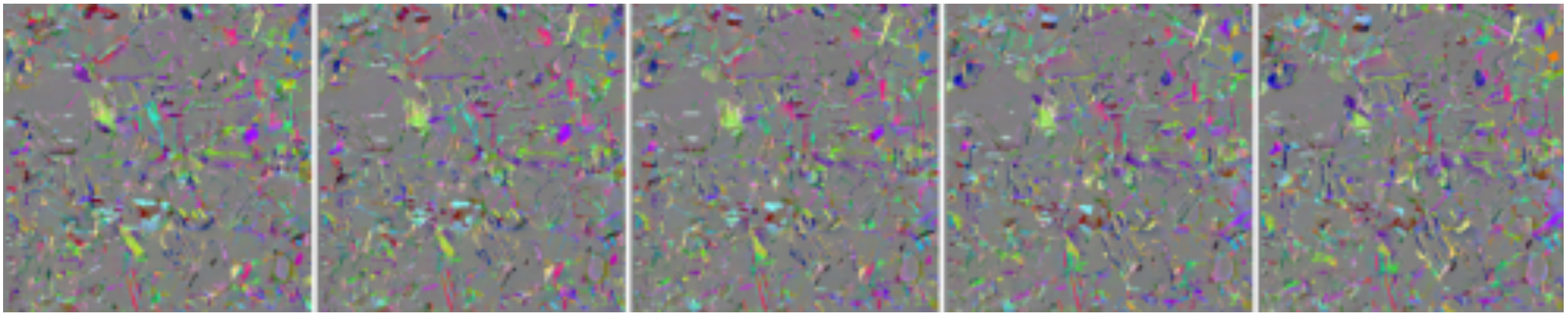}
         \caption{Sparse component produced by Baseline}
         \label{fig:ebsd_bayesian_s}
     \end{subfigure}
     \end{minipage}
        \caption{The tensor RPCA outputs on example EBSD slices. }
        \label{fig:ebsd}
\end{figure*}

We compare our SSL-based tensor RPCA method and the baseline (tuned by Optuna \cite{optuna_2019} using the loss \eqref{eq:l1loss}) on high-dimensional microscopy data from \cite{chapman2021afrl}. To provide context, data collection involved iteratively performing electron backscatter diffraction (EBSD) on a nickel-based superalloy. EBSD collects an image where the orientation of the local crystal, represented as Euler angles, is stored on each pixel. The superalloy was mechanically polished after each EBSD image collection, removing approximately 1 micron of material. This was repeated many times to obtain a sequence of slices containing orientation measurements, with an in-plane pixel spacing of 1 micron. The result is a volume of orientation information that can be considered as a high-dimensional tensor. See Fig.~\ref{fig:ebsd_y} for a visualization of consecutive slices. Note this dataset has no inherent notion of ground truth that we can perform supervised learning on for RPCA and contains only one sample. 

The motivation of applying RPCA here is that materials scientists usually arduously parse through these these slices manually. Based on Fig.~\ref{fig:ebsd_y}, changes from slice to slice are very subtle. Being able to separate the large slow moving sections from the sparse rapidly evolving parts could allow materials scientists to quickly identify critical rare events with implications on life-limiting behavior. 
Due to the dataset's significant amount of structure, we train our model for only 10 iterations of hyperparameter updates on $(250 \times 250 \times 250 \times 3)$ samples where we identify unique parameters for each sample using $\br = (250, 25, 250, 3)$ and $T=10$, again skipping updates along full rank modes. From Fig. \ref{fig:ebsd_ssl_x} and \ref{fig:ebsd_ssl_s}, our method extracts sparse components that capture detailed changes across slices and low-rank components that contain macro-scale patterns. We repeat the experiment for 50 iterations of the baseline method, with the results shown in Fig. \ref{fig:ebsd_bayesian_x} and \ref{fig:ebsd_bayesian_s}. Although the baseline resulted in similar loss values and captured similar quality low rank and sparse structures, it took many more iterations (about 5 times more) to obtain these results. Furthermore, we note that the per-iteration hyperparameter update runtimes of the baseline and our model were very similar, so our model also ran about 5 times faster. 

\vspace{-0.05in}
\section{Conclusions}
\label{sec:conclusion}
\vspace{-0.05in}
We have illustrated and empirically shown two ways deep unfolded SSL can be used to improve current approaches to tensor RPCA. First, a deep unfolded version of Algorithm \ref{alg:tensor_RPCA} with a new $\ell_1$-based loss function to train the hyperparameters can perform equivalently with supervised models and quickly provide satisfactory results in tasks where supervision is inapplicable. Second, self-supervised fine tuning improves recovery by finding a unique set of RPCA hyperparameters for each RPCA problem instance. Altogether, we have simultaneously extended the applicability of tensor RPCA to data-starved settings and enhanced its performance. Possible future adaptations include extending our method to design similar models for tensor algorithms that are centered around reconstruction like tensor completion. It would also be of interest to replace our recurrent layers with other architectures designed for sequential inputs, such as transformers.

\section*{Acknowledgments}
 
The work of H. Dong and Y. Chi is supported in part by the Air Force D3OM2S Center of Excellence under FA8650-19-2-5209, by ONR under N00014-19-1-2404, by the DARPA TRIAD program under Agreement No.~HR00112190099, by NSF under CCF-1901199 and ECCS-2126634, and by the Carnegie Mellon University Manufacturing Futures Initiative, made possible by the Richard King Mellon Foundation. 
 
\bibliographystyle{alphaabbr}
\bibliography{refs.bib,strings.bib}

\end{document}